\documentstyle[epsfig]{llncs}

\newenvironment{maliste}%
{ \begin{list}%
        {$\bullet$}%
        {\setlength{\labelwidth}{30pt}%
         \setlength{\leftmargin}{20pt}%
         \setlength{\itemsep}{\parsep}}}%
{ \end{list} }

\begin{document}
\title{Multiagent Approach for the Representation 
       of Information in a Decision Support System}

\author{Fahem Kebair and Fr\'ed\'eric Serin}

\institute{Universit\'e du Havre, LITIS - Laboratoire d'Informatique, \\
           de Traitement de l'Information et des Syst\`emes,\\
           25 rue Philippe Lebon, 76058, Le Havre Cedex, France \\
	   \{fahem.kebair, frederic.serin\}@univ-lehavre.fr}

\maketitle

\begin{abstract} 
	In an emergency situation, the actors need an assistance allowing them to react 
	swiftly and efficiently. In this prospect, we present in this paper a decision 
	support system that aims to prepare actors in a crisis situation thanks to 
	a decision-making support. The global architecture of this system is presented 
	in the first part. Then we focus on a part of this system which is designed to 
	represent the information of the current situation. This part is composed of a 
	multiagent system that is made of factual agents. Each agent carries a semantic 
	feature and aims to represent a partial part of a situation. The agents develop 
	thanks to their interactions by comparing their semantic features using proximity 
	measures and according to specific ontologies. \\

\end{abstract}
\section{Introduction}
	Making a decision in a crisis situation is a complicated task. This is mainly due to the unpredictability and the rapid evolution of the environment state. 
	Indeed, in a critic situation time and resources are limited. Our knowledge about the environment is incomplete and uncertain, verily obsolete. 
	Consequently, it is difficult to act and to adapt to the hostile conditions of the world. This makes sense to the serious need of robust, dynamic and 
	intelligent planning system for search-and-rescue operations to cope with the changing situation and to best save people \cite{kitano01}. The role of 
	such a system is to provide an emergency planning that allows actors to react swiftly and efficiently to a crisis case. 

	In this context, our aim is to build a system designed to help decision-makers manage cases of crisis with an original representation of information. 
	From the system point of view, detecting a crisis implies its representation, its characterisation and its comparison permanently with other crisis  stored 
	in scenarios base. The result of this comparison is provided to the user as the answer of the global system .

	The idea began with the  speech interpretation of human actors during a crisis \cite{cardon97}, \cite{duran99}. The goal was to build an 
	\textit{information, and communication system} (ICS) which enables the management of emergency situations by interpreting aspects communications created 
	by the actors. Then, a \textit{preventive vigil system} (PVS) \cite{bouk02} was designed with the mean of some technologies used in the ICS modelling 
	as: semantic features, ontologies, and agents with internal variables and behavioural automata. The PVS aims either to prevent a crisis or to deal with it 
	with a main internal goal: detecting a crisis.

	Since 2003, the architecture of the PVS was redesigned with a new specificity, that is the generic aspect; generic is used here with different meaning from 
	\cite{woold98}. A part of the global system, which is responsible of the dynamic information representation of the current situation, was applied to the 
	game of Risk and tested thanks to a prototype implemented in Java \cite{patrick05}. However, we postulate that some parts of the architecture and, at a deeper 
	level, some parts of the agents were independent of the subject used as application. Therefore, the objective at present is to connect this part to the other 
	parts, that we present latter in this paper, and to test the whole system on various domains, as RoboCup Rescue \cite{robocup} and e-learning.
	
	We focus here on the modelling of the information representation part of the system that we intend to use it in a crisis management support system.

	The paper begins with the presentation of the global system architecture. The core of the system is constituted by a multiagent system (MAS) which is structured 
	on three multiagent layers. Then, in section 3, we explain the way we formalise the environment state and we extract information related to it, which are written in the form 
	of semantic features. The latter constitute data that feed the system permanently and that carry information about the current situation. 
	The semantic features are handled by factual agents and are compared the one with the other using specific ontologies \cite{bouk03}. 

	Factual agents, that compose the first layer of the core, are presented thereafter in section 4. Each agent carries a semantic feature and aims to reflect a partial 
	part of the situation. We present their structures and their behaviours inside their organisation using internal automaton and indicators.

	Finally, we present a short view about the game of Risk test in which we describe the model application and the behaviour of factual agents.

\section{Architecture of the Decision Support System}
	The role of the \textit{decision support system} (DSS) is to provide a decision-making support to the actors in order to assist them during a crisis case. 
	The DSS allows also managers to anticipate the occur of potential incidents thanks to a dynamic and a continuous evaluation of the current situation.
	Evaluation is realised by comparing the current situation with past situations stored in a scenarios base. The latter can be viewed as one part of the knowledge 
	we have on the specific domain. 

	The DSS is composed of a core and three parts which are connected to it (figure 1):

	\begin{maliste}
	   \item A set of user-computer interfaces and an intelligent interface allow the core to communicate with the environment. The intelligent interface controls and manages 
	the access to the core of the authenticated users, filters entries information and provides actors with results emitted by the system;
	   \item An \textit{inside query MAS} ensures the interaction between the core and world information. These information represent the knowledge the core need. The knowledge includes 
	the scenarios, that are stored in a scenarios base, the ontologies of the domain and the proximity measures;
	   \item An \textit{outside query MAS} has as role to provide the core with information, that are stored in network distributed information systems.
	\end{maliste}
	
	\begin{figure}
 	   \centering
 	   \epsfig{figure=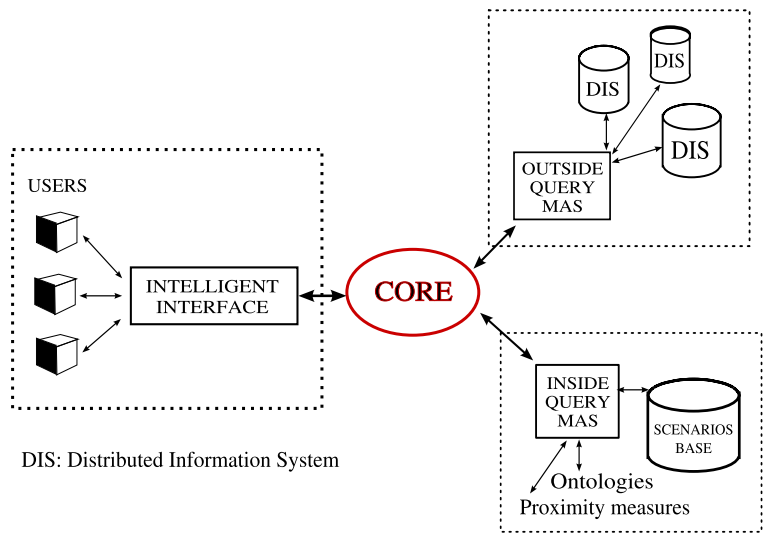,width=10cm,height=7cm,angle=0} 
 	   \caption{General Architecture of the DSS}
 	\end{figure}
	
	The core of the decision support system is made of a MAS which is structured on three layers. The latter contain specific agents that differ in their objectives and their 
	communications way. In a first time, the system describes the semantic of the current situation thanks to data collected from the environment. Then it analyses pertinent 
	information extracted from the scenario. Finally, it provides an evaluation of the current situation and a decision support using a dynamic and incremental case-base reasoning. 

	The three layers of the core are:

	\begin{maliste}
	   \item The lowest layer: factual agents;
	   \item The intermediate layer: synthesis agents;
	   \item The highest layer: prediction agents.
	\end{maliste}

	Information are coming from the environment in the form of semantic features without a priori knowledge of their importance. The role of the first layer (the lowest one) is to 
	deal with these data thanks to factual agents and let emergence detect some subsets of all the information \cite{galinho03}. More precisely, the set of these agents will enable the 
	appearance of a global behaviour thanks to their interactions and their individual operations. The system will extract thereafter from this behaviour the pertinent information that 
	represent the salient facts of the situation. 

	The role of the \textit{synthesis agents} is to deal with the agents emerged from the first layer. Synthesis agents aim to create dynamically factual agents clusters according to their evolutions. 
	Each cluster represents an observed scenario. The set of these scenarios will be compared to past ones in order to deduce their potential consequences.
	 
	Finally, the upper layer, will build a continuous and incremental process of recollection for dynamic situations. This layer is composed of \textit{prediction agents} and has as goal to evaluate 
	the degree of resemblance between the current situation and its associate scenario continuously. Each prediction agent will be associated to a scenario that will bring it closer, from 
	semantic point of view, to other scenarios for which we know already the consequences. The result of this comparison constitutes a support information that can help a manager to make a good decision.
 
	\begin{figure}
 	   \centering
 	   \epsfig{figure=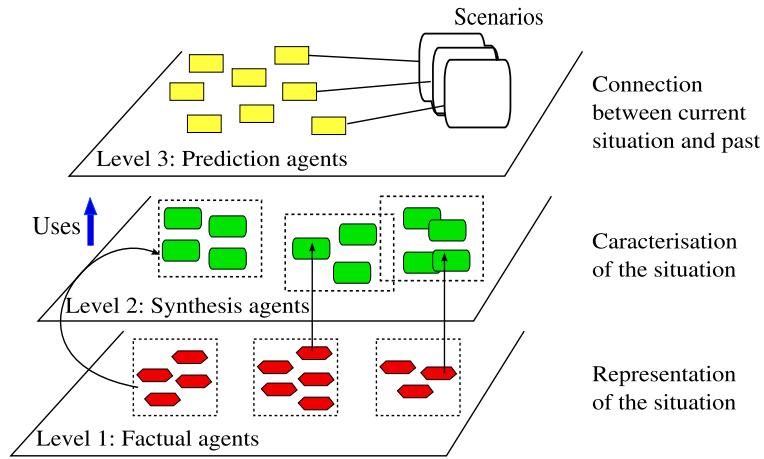,width=10cm,height=6cm,angle=0} 
 	   \caption{Architecture of the Core}
 	\end{figure}

\section{Environment Study and Creation of Semantic Features}
   \subsection{Situation Formalisation}
	To formalise a situation means to create a formal system, in an attempt to capture the essential features of the real-world. To realise this, we model the world as 
	a collection of objects, where each one holds some properties. The aim is to define the environment objects following the object paradigm. Therefore, we build a structural 
	and hierarchical form in order to give a meaning to the various relations that may exist between them. The dynamic change of these objects states and more still the interactions 
	that could be entrenched between them will provide us a snapshot description of the environment. In our context, information are decomposed in atomic data 
	where each one is associated to a given object. 
   \subsection{Semantic Features}
	A semantic feature is an elementary piece of information coming from the environment and which represents a fact that occurred in the world. Each semantic feature is related to an object 
	(defined in section 3.1), and allows to define all or a part of this object. A semantic feature has the following form: 
	(key, $(qualification, value)^{+}$), where key is the described object and $(qualification, $ 
	$value)^{+}$ is a set of couples formed by: the qualification of the object and its associated value. As example of a semantic feature related to a phenomenon object: 
	(phenomenon$\#1$, type, fire, location, $\#$4510, time, 9:33). The object described by this semantic feature is phenomenon$\#1$, and has as qualifications: type, location, and time.

	The modelling of semantic features makes it possible to obtain a homogeneous structure. This homogeneity is of primary importance because it allows to establish comparisons between these data. 
	The latter are managed by factual agents, where each one carries one semantic feature and of which behaviour depends on the type of this information.

	According to FIPA communicative acts \cite{FIPA}, the agents must share the same language and vocabulary to communicate. The use of semantic features in communications process implies to define an ontology.

	Inside the representation layer (the first layer of the system), agents evolve by comparing their semantic features. These comparisons allow to establish semantic distances between the agents, and 
	are computed thanks to proximity measures. 
	
	We distinguish three types of proximities: time proximity, spatial proximity and semantic proximity. The global proximity multiplies these three proximities together. 
	The measurement of a semantic proximity is related to ontologies. Whereas time proximity and spatial proximity are computed according to specific functions.

	Proximities computation provides values on $[-1,1]$ and is associated to a scale. The reference value in this scale is 0 that means neutral relation between the two compared semantic features. 
	Otherwise, we can define the scale as follow: 0.4=Quiet Close, 0.7=Close, 0.9=Very Close, 1=Equal. Negative values mirrors positive ones (replacing close by different).

\section{Factual Agents}
   \subsection{Presentation and Structure of a Factual Agent}
	Factual agents are hybrid agents, they are both cognitive and reactive agents. They have therefore the following characteristics: reactivity, proactiveness and social
	ability \cite{woold02}. Such an agent represents a feature with a semantic character and has also to formulate this character feature, a behaviour \cite{cardon04}. This 
	behaviour ensures the agent activity, proactiveness and communication functions.

	The role of a factual agent is to manage the semantic feature that it carries inside the MAS. The agent must develop to acquire a dominating place in its organisation and consequently, to make 
	prevail the semantic category which it represents. For this, the factual agent is designed with an implicit goal that is to gather around it as much friends as possible in order to build 
	a cluster. In other words, the purpose of the agent is to add permanently in its acquaintances network a great number of semantically close agents. The cluster formed by these agents is recognized by the system 
	as a scenario of the current situation and for which it can bring a potential consequence. A cluster is formed only when its agents are enough strong and consequently they are in an advanced state in 
	their automaton. Therefore, the goal of the factual agent is to reach the action state, in which is supreme and its information may be regarded by the system as relevant. 

	\begin{figure}
	   \centering
	   \epsfig{figure=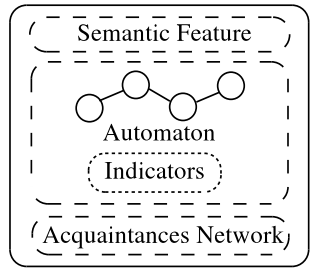,width=4cm,height=3.5cm, angle=0}  
	   \caption{Structure of a Factual Agent}
	\end{figure} 

	An internal automaton describes the behaviour and defines the actions of the agent. Some indicators and an acquaintances network allow the automaton operation, that means they help the agent to progress 
	inside its automaton and to execute actions in order to reach its goal. These characteristics express the proactiveness of the agent. 

	The acquaintances network contains the addresses of the friends agents and the enemies agents used to send messages. This network is dynamically constructed and permanently updated. Agents are friends 
	(enemies) if their semantic proximities are strictly positive (negative).

   \subsection{Factual Agent Behaviour}

\subsubsection{Behavioural Automaton} 
 
	The internal behaviour of a factual agent is described by a generic augmented transition network (ATN). The ATN is made of four states \cite{cardon97} (quoted above) linked by 
	transitions: 

	\begin{maliste}
	   \item \textit{Initialisation} state: the agent is created and enters in activities;
	   \item \textit{Deliberation} state: the agent searches in its acquaintances allies in order to achieve its goals;
	   \item \textit{Decision} state: the agent try to control its enemies to be reinforced;
	   \item \textit{Action} state: it is the state-goal of the factual agent, in which the latter demonstrates its strength by acting and liquidating its enemies.
	\end{maliste}

	\begin{figure}
 	   \centering
 	   \epsfig{figure=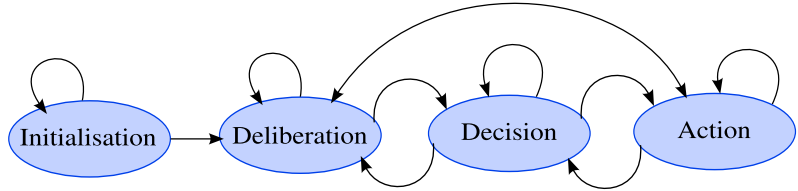,width=10.5cm,height=2.5cm, angle=0}  
 	   \caption{Generic Automaton of a Factual Agent}
 	\end{figure} 

	ATN transitions are stamped by a set of conditions and a sequence of actions. Conditions are defined as thresholds using internal indicators. The agent must validate thus one of its outgoing current state 
	transitions in order to pass to the next state. The actions of the agents may be an enemy aggression or a friend help. The choice of the actions to perform depend both on the type of the agent and its position in the ATN. 

      \subsubsection{Factual Agent Indicators}

	The dynamic measurement of an agent behaviour and its state progression at a given time are given thanks to indicators. These characters are significant parameters 
	that describe the activities variations of each agent and its structural evolution. In other words, the agent state is specified by the set of these significant 
	characters that allow both the description of its current situation and the prediction of its future behaviour \cite{cardon04} (quoted above). 

	Factual agent has five indicators, which are pseudoPosition (PP), pseudoSpeed (PS), pseudoAcceleration (PA), satisfactory indicator (SI) and constancy indicator (CI) \cite{galinho06}. 
	The ``pseudo'' prefix means that these indicators are not a real mathematical speed or acceleration: we chose a constant interval of time of one between two evolutions of 
	semantic features. PP represents the current position of an agent in the agent representation space. PS evaluates the PP evolution speed and PA means the PS evolution estimation.
	SI is a valuation of the success of a factual agent in reaching and staying in the deliberation state. This indicator measures the satisfaction degree of the agent. 
	Whereas, CI represents the tendency of a given factual agent to transit both from a state to a different state and from a state to the same state. This allows 
	the stability measurement of the agent behaviour.

	The compute of these indicators is according to this formulae where \textit{valProximity} depends on the category of a given application factual agents: 

	\begin{flushleft}
	\hspace{4cm}\textit{$PP_{t+1}$ = valPoximity} \\
	\hspace{4cm}\textit{$PS_{t+1}$ = $PP_{t+1}$ $-$ $PP_{t}$} \\
	\hspace{4cm}\textit{$PA_{t+1}$ = $PS_{t+1}$ $-$ $PS_{t}$} 
	\end{flushleft}

	PP, PS and PA represent thresholds that define the conditions of the ATN transitions. The definition of this conditions are specified to a 
	given application. As shown in the previous formulae, only PP is specific. However, PS and PA are generic and are deduced from PP. SI and CI are also independent of 
	the studied domain and are computed according to the agent movement in its ATN.

\section{Game of Risk Use Case}
	The first layer model has been tested on the game of Risk. We chose this game as application not only because it is well suited for crisis management but 
 	also we apprehend the elements and the actions on such an environment. Moreover we have an expert \cite{galinho06} (quoted above) in our team who is able to evaluate and validate results at any moment. 

	As result, this test proved that this model allows the dynamic information representation of the current situation thanks to factual agents organisation. Moreover we 
	could study the behaviour and the dynamic evolution of these agents.

	Risk is a strategic game which is composed of a playing board representing a map of forty-two territories that are distributed on six continents. A player wins by conquering all 
	territories or by completing his secret mission. In turn, each player receives and places new armies and may attack adjacent territories. An attack is one or more battles fought 
	with dice. Rules, tricks and strategies are detailed in \cite{risk}.
	
	\begin{figure}
 	   \centering
 	   \epsfig{figure=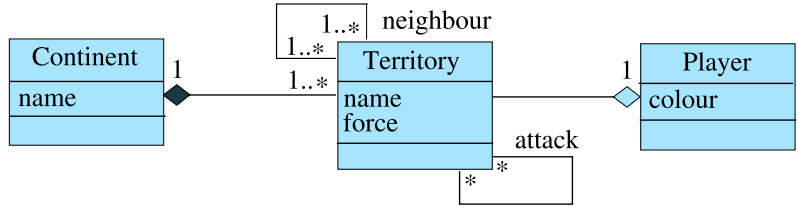,width=10.5cm,height=2.70 cm, angle=0}  
 	   \caption{Class Diagram for the Game of Risk Representation}
 	\end{figure} 

	The representation layer of the system has as role to simulate the game unwinding and to provide a semantic instantaneous description of its current state. To achieve this 
	task, we began by identifying the different objects that define the game board (figure 5) and which are: territory, player, army and continent. Continents and territories are
	regarded as descriptions of a persistent situation. Whereas, armies and players are activities respectively observed (occupying a territory) and driving the actions. 
	
	From this model we distinguish two different types of semantic features: a player type and a territory type. For example (Quebec, player, green, nbArmies, 4, time, 4) 
	is a territory semantic feature that means Quebec territory is owned by the green player and has four armies. However, (blue, nbTerritories, 4, time, 1) is a player semantic 
	feature that signifies a blue player has four territories at step 1. 

	The first extracted semantic features of the initial state of the game cause the creation of factual agents. For example, a semantic feature as 
	(red, nbTerritories, 0, time, 1) will cause the creation of red player factual agent.  

	During the game progression, the entry of a new semantic feature to the system may affect some agents state. A factual agent of type (Alaska, player, red, nbArmies, 3, time, 10) 
	become (Alaska, player, red, nbArmies, -2, time, 49) with the entry of the semantic feature (Alaska, player, red, nbArmies, 1, time, 49). Alaska agent sends messages containing its semantic 
	feature to all the other factual agents to inform them about its change. The other agents compare their own information with the received one. If an agent is interested by this message 
	(the proximity measure between the two semantic features is not null) it updates its semantic feature accordingly. If the red player owned GB before the semantic feature
	(GB, player, blue, nbArmies, 5, time, 52), both red player and blue player will receive messages because of the change of the territory owner.

	If we take again the preceding example (Alaska territory), Alaska agent computes its new PP (valProximity). The computation of valProximity in our case is given by: number of armies (t) 
	- number of armies (t-1) e.g. here valProximity = 1-3 = -2. PS and PA are deduced thereafter from PP.
	The agent verify then the predicates of its current state outgoing transitions in order to change state. To pass from \textit{Deliberation} state to \textit{Decision} state for 
	example the PS must be strictly positive. During this transition, the agent will send a \textit{SupportMessage} to a friend and an \textit{AgressionMessage} to an enemy.

%
\section{Conclusion}

	The paper has presented a decision support system which aims to help decision-makers to analyse and evaluate a current situation. The core of the system rests on an        
	agent-oriented multilayer architecture. We have described here the first layer which aims to provide a dynamic information representation of the current situation 
	and its evolution in time. This part is modelled with an original information representation methodology which is based on the handle of semantic features
	using a factual agents organisation.

	The model of the first layer was applied on the game of Risk. Results provided by this test correspond to our attempts, which consist on the dynamic representation of information. 
	This application allowed us to track the behaviour of factual agents and to understand their parameters which are the most accurate to characterise information. Moreover, we consider that a 
	great part of the system is generic and may be carried into other fields. Currently, we intend in a first time to connect the representation layer to the two other and to apply thereafter 
	the whole system on more significant domains as RoboCup Rescue and e-learning.

\end{document}